\documentclass[conference]{IEEEtran}
\IEEEoverridecommandlockouts

\usepackage{cite}
\usepackage{amsmath,amssymb,amsfonts}
\usepackage{algorithmic}
\usepackage{graphicx}
\usepackage{textcomp}
\usepackage{hyperref}
\usepackage{enumitem}
\usepackage{amsmath,amssymb}
\usepackage{graphicx,verbatim}
\usepackage{booktabs}
\usepackage[table,xcdraw]{xcolor} 
\usepackage[normalem]{ulem}
\useunder{\uline}{\ul}{}
\usepackage{multirow} 
\usepackage{bbding}
\usepackage{caption}
\usepackage{adjustbox}

\usepackage{hyperref}
\hypersetup{  colorlinks=true, linkcolor=blue, 
  citecolor=blue, urlcolor=blue, pdfborder={0 0 0} 
}

\def\BibTeX{{\rm B\kern-.05em{\sc i\kern-.025em b}\kern-.08em
    T\kern-.1667em\lower.7ex\hbox{E}\kern-.125emX}}
\begin{document}

\title{Cross-Stain Contrastive Learning for Paired Immunohistochemistry and Histopathology Slide Representation Learning\\
\author{
\IEEEauthorblockN{Yizhi Zhang\textsuperscript{+}}
\IEEEauthorblockA{\textit{Communication University of China}\\
Beijing, China\\
yizhi.zhang@mails.cuc.edu.cn}
\and
\IEEEauthorblockN{Lei Fan\textsuperscript{+}}
\IEEEauthorblockA{\textit{UNSW Sydney}\\
Sydney, Australia\\
lei.fan1@unsw.edu.au}
\and
\IEEEauthorblockN{Zhulin Tao\textsuperscript{*}}
\IEEEauthorblockA{\textit{Communication University of China}\\
Beijing, China\\
taozl@cuc.edu.cn}
\and
\IEEEauthorblockN{Donglin Di}
\IEEEauthorblockA{\textit{Tsinghua University}\\
Beijing, China\\
donglin.ddl@gmail.com}
\and
\IEEEauthorblockN{Yang Song}
\IEEEauthorblockA{\textit{UNSW Sydney}\\
Sydney, Australia\\
yang.song1@unsw.edu.au}
\and
\IEEEauthorblockN{Sidong Liu}
\IEEEauthorblockA{\textit{Macquarie University}\\
Sydney, Australia\\
sidong.liu@mq.edu.au}
\and
\IEEEauthorblockN{Cong Cong\textsuperscript{*}}
\IEEEauthorblockA{\textit{Macquarie University}\\
Sydney, Australia\\
thomas.cong@mq.edu.au}
\thanks{\textsuperscript{+}Equal contribution. \textsuperscript{*}Corresponding authors.}
}


}

\maketitle

\begin{abstract}
Universal, transferable whole-slide image (WSI) representations are central to computational pathology. 
Incorporating multiple markers (\textit{e.g.}, immunohistochemistry, IHC) alongside H\&E enriches H\&E-based features with diverse, biologically meaningful information.
However, progress is limited by the scarcity of well-aligned multi-stain datasets. Inter-stain Misalignment shifts corresponding tissue across slides, hindering consistent patch-level features and degrading slide-level embeddings.
To address this, we curated a slide-level aligned, five-stain dataset (H\&E, HER2, KI67, ER, PGR) to enable paired H\&E–IHC learning and robust cross-stain representation.
Leveraging this dataset, we propose Cross-Stain Contrastive Learning (CSCL), a two-stage pretraining framework:
a lightweight adapter trained with patch-wise contrastive alignment to improve the compatibility of H\&E features with corresponding IHC-derived contextual cues;
and slide-level representation learning with Multiple Instance Learning (MIL), which uses a cross-stain attention fusion module to integrate stain-specific patch features and a cross-stain global alignment module to enforce consistency among slide-level embeddings across different stains.
Experiments on cancer subtype classification, IHC biomarker status classification, and survival prediction, show consistent gains by yielding high-quality, transferable H\&E slide-level representations.
The code and data are available at: \url{https://github.com/lily-zyz/CSCL}.
\end{abstract}


\section{Introduction}
Whole-slide image (WSI) representation learning is fundamental to computational pathology (CPath). However, gigapixel slides pose challenges. To address this, existing frameworks \cite{song2023artificial,fan2022cancer,tang2025prototype} patchify slides and aggregate with Multiple Instance Learning (MIL) \cite{ilse2018attention,shao2021transmil}.
Although effective, most MIL methods remain intra-modal embeddings that come only from the same slide and stain, limiting diversity and expressiveness of the resulting slide-level features \cite{qu2025multimodal,fan2021learning}.
Recently, MADELEINE \cite{jaume2024multistain} enhances slide-level representation learning via global–local cross-stain alignment, but the absence of patch-level alignment forces coarse matching and mispairs H\&E with unrelated IHC patches, degrading representations \cite{liu2022bci}.
It also concatenates multi-stain features without modeling inter-dependencies, overlooking the contextual relationships among stain-specific embeddings of the same region.
Overall, current multi-stain approaches either miss fine-grained interactions or impose rigid stain availability, reducing generalizability and clinical applicability.

Accordingly, we propose Cross-Stain Contrastive Learning (CSCL), a pretraining framework for enhanced slide-level representations. To address stain misalignment, a paired five-stain dataset (H\&E, HER2, KI67, ER, PGR) is curated. Leveraging this dataset, CSCL operates in two stages to effectively leverage cross-stain information:
To enhance patch-level feature extraction, a lightweight adapter placed after the H\&E-pretrained encoder \cite{lu2024visual} is fine-tuned via Cross-Stain Patch-wise Alignment, leveraging the paired dataset to align H\&E features with contextual cues from corresponding IHC patches and improve cross-stain compatibility. 
Building on the enriched patch features, MIL-based slide-level representation learning employs two components: Cross-Stain Attention Fusion to capture same-region inter-stain interactions, and Cross-Stain Global Alignment to align slide-level embeddings across stains.
This yields robust, generalizable representations, enabling the MIL model to accommodate variation in staining protocols and institutional sources. Through cross-stain contrastive pretraining, CSCL learns universal, transferable H\&E slide-level representations, improving adaptability across protocols and clinical settings. 

The main contributions include: 1) CSCL is supported by an aligned multi-stain dataset pairing H\&E with four IHC WSIs (HER2, KI67, ER, PGR), to our knowledge, the first aligned resource tailored for slide representation learning; 2) Patch-wise alignment supervises a lightweight adapter, enhancing H\&E feature extraction with contextual cues from corresponding IHC patches; and 3) Comprehensive experiments on cancer subtype classification, IHC biomarker status classification, and survival prediction show consistent gains.

\section{RELATED WORK}
\textbf{Multi-stain Dataset and Pretraining.} ACROBAT \cite{weitz2024acrobat} offers paired slides of five routine stains, but lacks patch-level alignment across stains, leading to inconsistent representations and hindering fine-grained learning.
To address this, repeated staining on the same section can ensure patch-level alignment \cite{kataria2023automating}, yet risks tissue degradation and registration artifacts, while synthetic generation \cite{brazdil2022automated} may underrepresent real-world variability. Moreover, many datasets \cite{kataria2023automating,brazdil2022automated} are not publicly accessible, limiting reproducibility and broader adoption.

Multi-stain representation learning captures complementary cross-stain cues for richer tissue characteristics. MADELEINE \cite{jaume2024multistain} uses Graph Optimal Transport for local alignment, but absent spatial co-registration can mispair patches \cite{liu2022bci}. 
In contrast, we build on a patch-aligned multi-stain dataset and adopt interaction-aware pretraining to learn flexible, generalizable cross-stain representations.

\textbf{Histopathology Image Representation Learning.}
Given scarce pixel- or region-level annotations, self-supervised learning (SSL) predominates, typically (1) learning meaningful patch representations~\cite{fan2024patch} and (2) aggregating them into robust slide-level embeddings. Large-scale visual encoders can extract informative patch embeddings \cite{azizi2023robust,chen2024towards,filiot2023scaling,fan2022fast,fan2025grainbrain}. In parallel, vision–language models leverage web-scale image–text pairs for semantically grounded patch features \cite{gamper2021multiple, huang2023visual}. However, both often treat patches independently and within a single stain, overlooking cross-stain signals.

Slide representation learning aggregates patch features into holistic WSI embeddings. Hierarchical slide pretraining encodes patches and then aggregates to slides, typically with contrastive or reconstruction objectives \cite{wang2025hypergraph,chen2022scaling}. Recent work augments slide-level representation learning with multi-modal supervision, \textit{e.g.}, transcriptomics \cite{jaume2024transcriptomics,qu2025memory}, adding biological or clinical context. Yet fine-grained, stain-specific context in co-registered multi-stain slides remains underexplored. Leveraging complementary cues from multiple stains at patch and slide levels, our method learns robust, transferable, biologically grounded WSI representations.

\begin{figure*}[h!]
    \centering
    \includegraphics[width=1\textwidth]{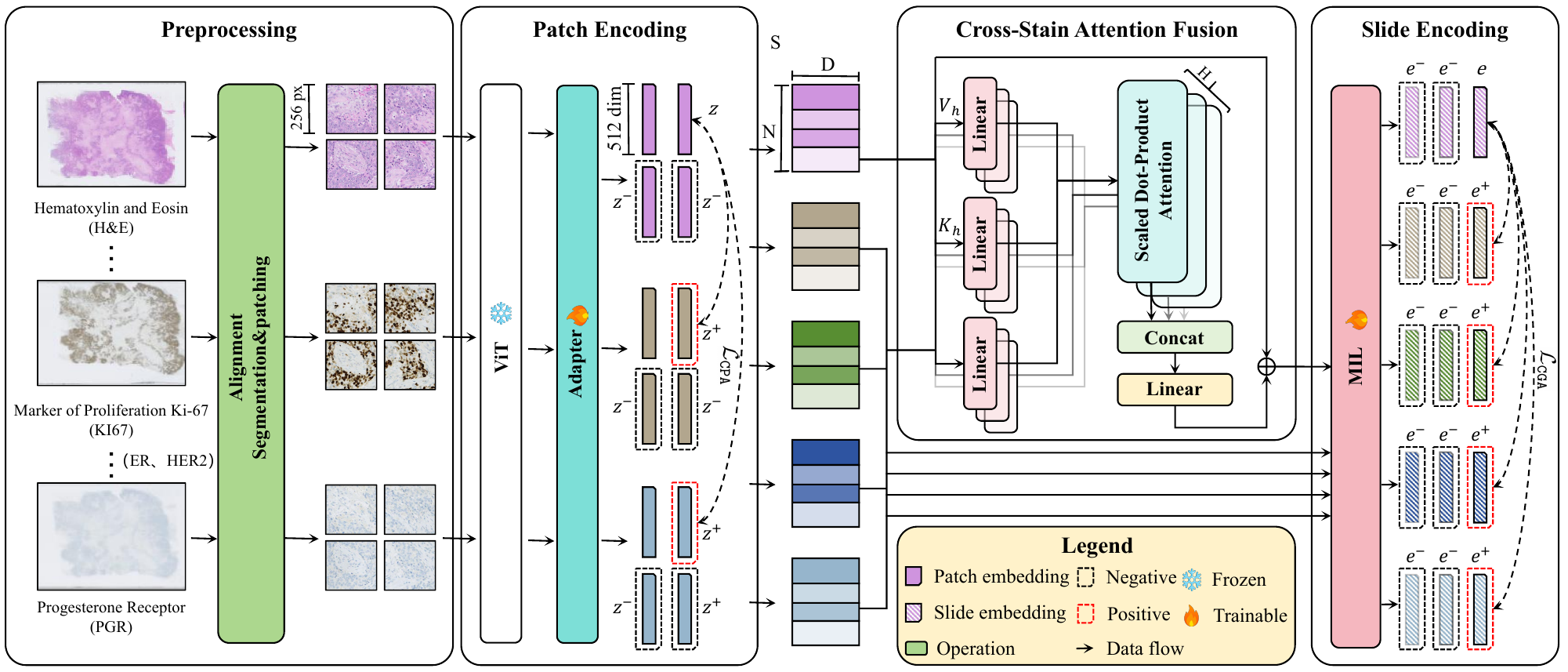}
    \caption{\textbf{Overview of CSCL.} \textbf{Preprocessing}: Five types of stained WSIs are aligned, followed by tissue segmentation and 256$\times$256 patching. \textbf{Patch Encoding}: Patches are fed into a Vision Transformer (ViT); an adapter optimized by $\mathcal{L}_{\mathtt{CPA}}$ refines them to enforce consistent cross-stain patch-level alignment across different stains. \textbf{Cross-Stain Attention Fusion}: Multi-stain features are fused to enrich representations. \textbf{Slide Encoding}: Multi-stain features are fused to enrich H\&E representations. \textbf{Slide Encoding}: For each stain, patch embeddings are aggregated by MIL into stain-specific slide embeddings; subsequently, $\mathcal{L}_{\mathtt{CGA}}$ aligns H\&E with IHC, yielding consistent, informative, and stain-invariant slide features.}
    \label{fig:Overview}
\end{figure*}

\section{Method}
We propose Cross-Stain Contrastive Learning (CSCL), a pretraining framework that leverages paired multi-stain information to learn informative, transferable slide-level features from H\&E WSIs. The overall pipeline of CSCL is illustrated in Fig.~\ref{fig:Overview}.

\subsection{Paired Multi-stain WSI Dataset Construction}
\label{sec:pre}
A paired multi-stain dataset was curated from ACROBAT \cite{rantalainen2023acrobat}, which provides WSIs under five staining protocols. ACROBAT’s WSI quality varies substantially: some slides show contamination (\textit{e.g.}, dust) and tissue overlap, while others exhibit low resolution due to imaging instrument limitations.
To ensure slide-level representation quality, lower-quality slides were excluded, yielding a curated subset of 171 WSI pairs, each pair corresponding to the same tissue sample under different stains.
Spatial consistency across stains is achieved with an automated registration algorithm \cite{wodzinski2024regwsi} that aligns WSIs across different staining protocols. The registration process comprises three stages:
\begin{enumerate}
\item \textbf{Preprocessing}: Original WSIs are converted to grayscale, normalized, and enhanced with CLAHE \cite{pizer1990contrast} to improve contrast. Gaussian smoothing is applied, and images are resampled to ensure uniform resolution.
\item \textbf{Initial Alignment}: Coarse tissue alignment with pre-trained SuperPoint~\cite{detone2018superpoint} and SuperGlue~\cite{sarlin2020superglue} yields rough spatial correspondence between WSIs.
\item \textbf{Nonrigid Registration}: A refined alignment employing affine transformations is carried out to achieve precise spatial correspondence. This nonrigid step improves the matching of tissue structures across WSIs.
\end{enumerate}
After registration, non-overlapping \(256\times256\) patches at \(10\times\) were extracted from each WSI, yielding spatially aligned, stain-paired inputs for downstream encoding. 

\subsection{Cross-Stain Contrastive Learning}
\label{sec:stain}
CSCL comprises three components aligned with the framework stages:
(1) Cross-Stain Patch-wise Alignment ($\mathtt{CPA}$) for adapter-based patch-level adaptation,
(2) Cross-Stain Attention Fusion ($\mathtt{CAF}$) for integrating stain-specific patch features, and
(3) Cross-Stain Global Alignment ($\mathtt{CGA}$) for aligning slide-level embeddings across stains.

\subsubsection{$\mathtt{CPA}$}
\label{sec:local}
To enhance patch-level feature extraction in the multi-stain setting, a lightweight adapter is placed after a frozen H\&E-pretrained encoder \cite{lu2024visual}.
The adapter adapts the encoder by aligning H\&E-derived features with contextual cues from corresponding IHC stains and is optimized with the $\mathtt{CPA}$ contrastive objective. 
Despite registration, perfect pixel-level correspondence is often unattainable due to staining variation, tissue distortion, or sectioning artifacts.
$\mathtt{CPA}$ promotes semantic alignment of corresponding patches while remaining robust to minor misalignments.
Given $M$ stains comprising one H\&E and $C$ IHC stains, for an H\&E-stained patch $z$, we denote $z_c^{+}$ as the corresponding IHC patch (\emph{i.e.}, spatially aligned patch) in stain $c$ $(c \in \{1,\dots,C\})$, and $z^{-}$ as the set of non-corresponding patches across all stains.
The $\mathtt{CPA}$ loss is defined as:
\begin{equation}
\label{eq:cpa}
\mathcal{L}_{\mathtt{CPA}} = \mathbb{E}{z \sim Z} \sum_{c=1}^{C} \frac{w_t(z, z_c^{+})}{W_c^t} \cdot \mathcal{L}_{\text{InfoNCE}}(z, z_c^{+}, z^{-}),
\end{equation}
where $t$ denotes the current iteration, and $W_c^t = \sum_{z} w_t(z, z_c^{+})$, with the adaptive weighting function $w_t$ defined as, $w_t(z,z^{+}) = \bigl(1-g(\tfrac{t}{T})\bigr) + g\bigl(\tfrac{t}{T}\bigr) \times h(z\cdot z^{+})$. Here, $T$ is the total number of training iterations, $h(\cdot)$ is a function of the anchor-positive cosine similarity, and $g(\cdot)$ is a scheduling function. Moreover, the InfoNCE loss is defined as:
\begin{equation}
\mathcal{L}_{\text{InfoNCE}}(z, z_c^{+}, z^{-}) = -\log \frac{\exp(\frac{z \cdot z_c^{+}}{\tau})}{\exp(\frac{z \cdot z_c^{+}}{\tau}) + \sum_{n=1}^{N} \exp(\frac{z \cdot z_n^{-}}{\tau})},
\end{equation}
with $\tau$ represents a temperature parameter and $N$ denotes the number of patches within a slide.

\begin{table*}[t]
\centering
\caption{Comparison of AUC performance across multiple CPath frameworks on biomarker classification, reported for $k=1$, $k=5$, $k=10$, and $k=25$. Reported scores are the average AUC over ER, PR, and HER2 status prediction.}
\label{tab:comparison_IHCclassification}
\resizebox{\textwidth}{!}{
\begin{tabular}{c|cccc|cccc}
\toprule
\multirow{2}[0]{*}{Model} & \multicolumn{4}{c}{BCNB ($\uparrow$)} & \multicolumn{4}{c}{BRCA ($\uparrow$)} \\
& k=1 & k=5 & k=10 & k=25 & k=1 & k=5 & k=10 & k=25\\
\midrule
TransMIL & 0.511\textsubscript{$\pm$0.047} & 0.558\textsubscript{$\pm$0.064} & 0.595\textsubscript{$\pm$0.039} & 0.627\textsubscript{$\pm$0.054} & 0.513\textsubscript{$\pm$0.033} & 0.594\textsubscript{$\pm$0.037} & 0.596\textsubscript{$\pm$0.022} & 0.645\textsubscript{$\pm$0.046} \\
DSMIL & 0.548\textsubscript{$\pm$0.023} & 0.641\textsubscript{$\pm$0.072} & 0.692\textsubscript{$\pm$0.050} & 0.737\textsubscript{$\pm$0.067} & 0.548\textsubscript{$\pm$0.052} & 0.607\textsubscript{$\pm$0.081} & 0.646\textsubscript{$\pm$0.075} & 0.698\textsubscript{$\pm$0.023} \\
ACMIL & 0.537\textsubscript{$\pm$0.041} & 0.654\textsubscript{$\pm$0.036} & 0.693\textsubscript{$\pm$0.024} & 0.748\textsubscript{$\pm$0.052} & 0.530\textsubscript{$\pm$0.015} & 0.616\textsubscript{$\pm$0.054} & 0.660\textsubscript{$\pm$0.024} & 0.708\textsubscript{$\pm$0.058} \\
FRMIL & 0.552\textsubscript{$\pm$0.050} & 0.655\textsubscript{$\pm$0.051} & 0.699\textsubscript{$\pm$0.062} & 0.749\textsubscript{$\pm$0.027} & 0.544\textsubscript{$\pm$0.056} & 0.620\textsubscript{$\pm$0.023} & 0.657\textsubscript{$\pm$0.062} & 0.709\textsubscript{$\pm$0.025} \\
ABMIL & 0.543\textsubscript{$\pm$0.033} & 0.635\textsubscript{$\pm$0.013} & 0.689\textsubscript{$\pm$0.083} & 0.739\textsubscript{$\pm$0.051} & 0.513\textsubscript{$\pm$0.022} & 0.607\textsubscript{$\pm$0.034} & 0.646\textsubscript{$\pm$0.029} & 0.705\textsubscript{$\pm$0.060} \\ \midrule
HIPT & 0.507\textsubscript{$\pm$0.047} & 0.531\textsubscript{$\pm$0.029} & 0.568\textsubscript{$\pm$0.034} & 0.605\textsubscript{$\pm$0.033} & 0.497\textsubscript{$\pm$0.032} & 0.573\textsubscript{$\pm$0.018} & 0.611\textsubscript{$\pm$0.053} & 0.666\textsubscript{$\pm$0.028} \\
GigaSSL & 0.553\textsubscript{$\pm$0.073} & 0.623\textsubscript{$\pm$0.071} & 0.679\textsubscript{$\pm$0.056} & 0.701\textsubscript{$\pm$0.023} & 0.523\textsubscript{$\pm$0.045} & 0.604\textsubscript{$\pm$0.071} & 0.637\textsubscript{$\pm$0.041} & 0.690\textsubscript{$\pm$0.047} \\
GigaPath & 0.522\textsubscript{$\pm$0.061} & 0.599\textsubscript{$\pm$0.042} & 0.634\textsubscript{$\pm$0.067} & 0.680\textsubscript{$\pm$0.056} & 0.502\textsubscript{$\pm$0.072} & 0.575\textsubscript{$\pm$0.047} & 0.617\textsubscript{$\pm$0.039} & 0.667\textsubscript{$\pm$0.063} \\
TANGLE & 0.554\textsubscript{$\pm$0.052} & 0.654\textsubscript{$\pm$0.056} & 0.686\textsubscript{$\pm$0.059} & 0.737\textsubscript{$\pm$0.019} & \underline{0.558\textsubscript{$\pm$0.019}} & 0.617\textsubscript{$\pm$0.027} & 0.649\textsubscript{$\pm$0.062} & 0.689\textsubscript{$\pm$0.029} \\ \midrule
MADELEINE & \underline{0.568\textsubscript{$\pm$0.056}} & \underline{0.678\textsubscript{$\pm$0.033}} & \underline{0.727\textsubscript{$\pm$0.017}} & \underline{0.765\textsubscript{$\pm$0.031}} & 0.551\textsubscript{$\pm$0.060} & \underline{0.621\textsubscript{$\pm$0.042}} & \underline{0.663\textsubscript{$\pm$0.024}} & \underline{0.711\textsubscript{$\pm$0.081}} \\
CSCL (Ours) & \textbf{0.598\textsubscript{$\pm$0.087}} & \textbf{0.681\textsubscript{$\pm$0.013}} & \textbf{0.729\textsubscript{$\pm$0.066}} & \textbf{0.767\textsubscript{$\pm$0.034}} & \textbf{0.567\textsubscript{$\pm$0.024}} & \textbf{0.654\textsubscript{$\pm$0.031}} & \textbf{0.690\textsubscript{$\pm$0.065}} & \textbf{0.712\textsubscript{$\pm$0.073}} \\
\bottomrule
\end{tabular}
}
\end{table*}

\subsubsection{$\mathtt{CAF}$}
\label{sec:scale}
Building on the stain-adapted embeddings from $\mathtt{CPA}$, $\mathtt{CAF}$ models inter-stain contextual relationships to enhance the H\&E representation. For each spatial location across $M$ co-registered stains, the corresponding patch embeddings are stacked into a tensor $S \in \mathbb{R}^{M \times N \times D}$, where $N$ is the number of patches and $D$ is the embedding dimension.

Cross-stain dependencies at each spatial location are modeled by permuting $S$ to $(N, M, D)$ and applying multi-head self-attention along the stain dimension, enabling selective integration of complementary IHC information into the H\&E embedding. For attention head $h$, the computation is:
\begin{equation}
\hat{S}_h = \text{softmax}\left(\frac{Q_h K_h^\top}{\sqrt{d_k}}\right)V_h,
\end{equation}
where $Q_h = W_h^Q S$, $K_h = W_h^K S$, $V_h = W_h^V S$, and $d_k$ is the dimensionality of keys per head. Outputs from all heads are concatenated and projected through a final learnable matrix, followed by a residual connection:
\begin{equation}
\acute{S} = \text{Concat}(\hat{S}_1, \dots, \hat{S}_H) W^O \oplus S.
\end{equation}
where $\acute{S}$ denotes the updated H\&E patch embedding.

The updated H\&E patch embeddings (from $\acute{S}$) replace the original H\&E features. The refined H\&E embeddings, together with the original features from the remaining IHC-stained patches, serve as inputs to MIL for subsequent slide-level representation learning.

\subsubsection{$\mathtt{CGA}$}
\label{sec:global}
Stain-invariant yet semantically consistent slide-level representations are promoted by  $\mathtt{CGA}$.
Using the $\mathtt{CAF}$-fused patch embeddings, a MIL model $f_{\theta}$ produces per-stain slide embeddings. For stain $m\in{1,\dots,M}$ with patch embeddings ${z_i^{(m)}}_{i=1}^N$, the slide-level embedding is:
\begin{equation}
e_m = f_{\theta}(\{z_i^{(m)}\}_{i=1}^N).
\end{equation}
With H\&E slide embedding $e$, same-tissue cross-stain embedding $e_c^{+}$, and negative embeddings $e^{-}$ from unrelated slides, the $\mathtt{CGA}$ loss is define as:
\begin{equation}
\label{eq:cga}
\mathcal{L}_{\mathtt{CGA}}(e, e_c^{+}, e^{-}) = -\log \frac{\exp\left(\frac{e \cdot e_c^{+}}{\tau}\right)}{\exp\left(\frac{e \cdot e_c^{+}}{\tau}\right) + \sum_{n=1}^{N} \exp\left(\frac{e \cdot e_n^{-}}{\tau}\right)},
\end{equation}
where $\tau$ is a temperature parameter. 

\subsection{Training and Inference}
Our training pipeline comprises two sequential stages.
In Stage 1, the adapter module is optimized with $\mathcal{L}_{\mathtt{CPA}}$ (Eq.~\ref{eq:cpa}), adapting H\&E-derived features to the multi-stain context. 
In Stage 2, the encoder and adapter are frozen; $\mathtt{CAF}$ fuses cross-stain patch features, and the MIL model $f_{\theta}$ is trained with $\mathcal{L}_{\mathtt{CGA}}$ (Eq.~\ref{eq:cga}).
At inference, only H\&E is used: patches are encoded by the frozen encoder–adaptor and aggregated by the trained $f_{\theta}$ to yield the slide-level representation. Cross-stain modules (attention fusion and contrastive alignment) are disabled, enabling single-stain operation without paired data.

\section{Experiments}
\subsection{Datasets and Experimental Details}
\subsubsection{Datasets}
Pretraining Dataset (Multi-stain): CSCL is trained on 171 curated sets of co-registered WSIs, each set comprising five stains: H\&E, PR, HER2, KI67, and ER. Downstream Evaluation Datasets: Molecular biomarker classification used H\&E-stained slides from BRCA~\cite{cancer2012comprehensive} for binary (positive vs.\ negative) ER (N=996), PR (N=993), and HER2 (N=693) status, and from BCNB~\cite{xu2021predicting} for ER/PR/HER2 status (N=933 each). Cancer subtype classification on BRCA~\cite{cancer2012comprehensive} distinguished IDC (N=542) vs.\ ILC (N=463). Survival prediction used H\&E slides with survival labels from BRCA (N=1{,}049), BLCA~\cite{tcga2014blca} (N=359), HNSC~\cite{cancer2015comprehensive} (N=392), and COADREAD~\cite{willett2013cancer} (N=296).

\subsubsection{Settings} 
ABMIL~\cite{ilse2018attention} served as the default MIL backbone and was trained with the adapter. The adapter+MIL produced slide embeddings, followed by a per-task linear classifier. Training used AdamW for 120 epochs (5 warm-up) with cosine LR decay \(10^{-4}\!\to\!10^{-8}\), batch size 24, on NVIDIA A800 GPUs. For comparability, tissue regions were segmented and non-overlapping \(256\times256\) patches at \(10\times\) were extracted. Both CSCL and baselines computed 512-dimensional patch embeddings using CONCH~\cite{lu2024visual}.
Importantly, multiple stains are required only during training; downstream evaluation uses H\&E only, thereby aligning with real-world clinical constraints and demonstrating the practical utility of our method. Evaluation employed \(k\)-shot AUC for molecular biomarker and subtype classification and 5-fold cross-validation with mean C-index for survival prediction.

\begin{table*}[t]
\centering
\begin{minipage}{0.60\textwidth}
\centering
\caption{Comparison of C-index performance across multiple CPath frameworks on survival prediction. Reported values are the average over 5-fold cross-validation. Best performance in \textbf{bold}, second best \underline{underlined}.}
\label{tab:comparison_survivalprediction}
\adjustbox{max width=\textwidth}{
\begin{tabular}{lccccc}
\toprule
Model & BRCA ($\uparrow$)    & BLCA ($\uparrow$)    & HNSC ($\uparrow$)   & COADREAD ($\uparrow$) & Avg ($\uparrow$) \\ \midrule
TransMIL & 0.697\textsubscript{$\pm$0.046} & \underline{0.644}\textsubscript{$\pm$0.091} & \underline{0.672}\textsubscript{$\pm$0.056} & 0.765\textsubscript{$\pm$0.043} & 0.696 \\
DSMIL & 0.679\textsubscript{$\pm$0.018} & 0.643\textsubscript{$\pm$0.071} & 0.667\textsubscript{$\pm$0.018} & 0.743\textsubscript{$\pm$0.017} & 0.683 \\
ACMIL & 0.700\textsubscript{$\pm$0.021} & 0.637\textsubscript{$\pm$0.064} & 0.668\textsubscript{$\pm$0.071} & \textbf{0.788}\textsubscript{$\pm$0.019} & \underline{0.698} \\
FRMIL & 0.662\textsubscript{$\pm$0.047} & 0.656\textsubscript{$\pm$0.073} & 0.670\textsubscript{$\pm$0.064} & 0.740\textsubscript{$\pm$0.011} & 0.682 \\
ABMIL & 0.669\textsubscript{$\pm$0.073} & 0.637\textsubscript{$\pm$0.023} & 0.672\textsubscript{$\pm$0.032} & 0.786\textsubscript{$\pm$0.063} & 0.691 \\ \midrule
HIPT & 0.547\textsubscript{$\pm$0.078} & 0.582\textsubscript{$\pm$0.059} & 0.593\textsubscript{$\pm$0.024} & 0.677\textsubscript{$\pm$0.047} & 0.600 \\
GigaSSL & 0.530\textsubscript{$\pm$0.038} & 0.546\textsubscript{$\pm$0.052} & 0.584\textsubscript{$\pm$0.063} & 0.669\textsubscript{$\pm$0.053} & 0.582 \\
GigaPath & 0.521\textsubscript{$\pm$0.081} & 0.535\textsubscript{$\pm$0.038} & 0.572\textsubscript{$\pm$0.035} & 0.658\textsubscript{$\pm$0.034} & 0.572 \\
TANGLE & 0.709\textsubscript{$\pm$0.028} & 0.637\textsubscript{$\pm$0.018} & 0.663\textsubscript{$\pm$0.043} & 0.753\textsubscript{$\pm$0.063} & 0.691 \\ \midrule
MADELEINE & \underline{0.715}\textsubscript{$\pm$0.041} & 0.635\textsubscript{$\pm$0.053} & 0.668\textsubscript{$\pm$0.074} & 0.759\textsubscript{$\pm$0.039} & 0.694 \\
CSCL (Ours) & \textbf{0.717}\textsubscript{$\pm$0.028} & \textbf{0.645}\textsubscript{$\pm$0.032} & \textbf{0.675}\textsubscript{$\pm$0.041} & \underline{0.768}\textsubscript{$\pm$0.032} & \textbf{0.701} \\
\bottomrule
\end{tabular}
}
\end{minipage}
\hfill
\begin{minipage}{0.38\textwidth}
\centering
\caption{Comparison of AUC performance across multiple CPath frameworks on breast cancer subtype classification (IDC vs. ILC). Best performance in \textbf{bold}, second best \underline{underlined}.}
\label{tab:comparison_classification}
\adjustbox{max width=\textwidth}{
\begin{tabular}{lccc}
\toprule
Model & k=1 ($\uparrow$)    & k=5 ($\uparrow$)    & k=10 ($\uparrow$) \\ \midrule
TransMIL & 0.555\textsubscript{$\pm$0.095} & 0.630\textsubscript{$\pm$0.091} & 0.639\textsubscript{$\pm$0.078} \\
DSMIL & 0.590\textsubscript{$\pm$0.034} & 0.718\textsubscript{$\pm$0.050} & 0.753\textsubscript{$\pm$0.073} \\
ACMIL & 0.603\textsubscript{$\pm$0.041} & 0.721\textsubscript{$\pm$0.101} & 0.748\textsubscript{$\pm$0.029} \\
FRMIL & 0.597\textsubscript{$\pm$0.016} & 0.723\textsubscript{$\pm$0.043} & 0.753\textsubscript{$\pm$0.034} \\
ABMIL & 0.574\textsubscript{$\pm$0.118} & 0.709\textsubscript{$\pm$0.038} & 0.738\textsubscript{$\pm$0.041} \\ \midrule
HIPT & 0.622\textsubscript{$\pm$0.039} & 0.693\textsubscript{$\pm$0.054} & 0.775\textsubscript{$\pm$0.039} \\
GigaSSL & 0.682\textsubscript{$\pm$0.066} & 0.787\textsubscript{$\pm$0.048} & 0.828\textsubscript{$\pm$0.042} \\
GigaPath & 0.587\textsubscript{$\pm$0.067} & 0.686\textsubscript{$\pm$0.049} & 0.754\textsubscript{$\pm$0.048} \\
TANGLE & 0.661\textsubscript{$\pm$0.123} & 0.819\textsubscript{$\pm$0.025} & 0.837\textsubscript{$\pm$0.032} \\ \midrule
MADELEINE & \underline{0.664}\textsubscript{$\pm$0.079} & \underline{0.858}\textsubscript{$\pm$0.032} & \underline{0.886}\textsubscript{$\pm$0.016} \\
CSCL (Ours) & \textbf{0.665}\textsubscript{$\pm$0.081} & \textbf{0.862}\textsubscript{$\pm$0.045} & \textbf{0.894}\textsubscript{$\pm$0.072} \\
\bottomrule
\end{tabular}
}
\end{minipage}
\end{table*}

\begin{table*}[t]
\centering
\caption{Ablation study of proposed modules. Best performance in \textbf{bold}, second best \underline{underlined}.}
\label{tab:Ablation_CPA_CGA_CFA}
\resizebox{\textwidth}{!}{
\small
\begin{tabular}{lccc|ccc|ccc|c}
\toprule
 & \multirow{2}{*}{$\mathtt{CPA}$} & \multirow{2}{*}{$\mathtt{CAF}$} & \multirow{2}{*}{$\mathtt{CGA}$} 
 & \multicolumn{3}{c|}{IHC Classification (BCNB) ($\uparrow$)} 
 & \multicolumn{3}{c|}{Cancer Subtyping (BRCA) ($\uparrow$)} 
 & \multicolumn{1}{c}{Survival} \\
 \cmidrule(lr){5-7} \cmidrule(lr){8-10}
& & & & k=1 & k=10 & k=25 & k=1 & k=10 & k=25 & (BRCA) ($\uparrow$) \\
\midrule
\multirow{6}{*}{\rotatebox{90}{TransMIL}} 
&\multicolumn{3}{c|}{\textit{Baseline}}  & 0.489$_{\pm0.027}$ & 0.535$_{\pm0.032}$ & 0.592$_{\pm0.053}$ & 0.599$_{\pm0.053}$ & 0.623$_{\pm0.024}$ & 0.728$_{\pm0.052}$ & 0.668$_{\pm0.084}$ \\
&  &  & \Checkmark & 0.512$_{\pm0.025}$ & 0.553$_{\pm0.028}$ & 0.612$_{\pm0.062}$ & 0.621$_{\pm0.017}$ & 0.664$_{\pm0.081}$ & 0.751$_{\pm0.034}$ & 0.671$_{\pm0.034}$ \\
& \Checkmark &  &  & 0.509$_{\pm0.054}$ & 0.541$_{\pm0.071}$ & 0.601$_{\pm0.035}$ & 0.614$_{\pm0.046}$ & 0.640$_{\pm0.043}$ & 0.749$_{\pm0.051}$ & 0.673$_{\pm0.039}$ \\
& \Checkmark & \Checkmark  &  & 0.528$_{\pm0.038}$ & \underline{0.585$_{\pm0.043}$} & \underline{0.633$_{\pm0.098}$} & \underline{0.635$_{\pm0.038}$} & \underline{0.693$_{\pm0.054}$} & \underline{0.768$_{\pm0.042}$} & 0.689$_{\pm0.056}$ \\
& \Checkmark  &  & \Checkmark  & \underline{0.544$_{\pm0.046}$} & 0.573$_{\pm0.062}$ & 0.628$_{\pm0.054}$ & 0.631$_{\pm0.064}$ & 0.688$_{\pm0.071}$ & 0.753$_{\pm0.069}$ & \underline{0.692$_{\pm0.049}$} \\
& \Checkmark & \Checkmark & \Checkmark & \textbf{0.564$_{\pm0.038}$} & \textbf{0.625$_{\pm0.076}$} & \textbf{0.667$_{\pm0.042}$} & \textbf{0.661$_{\pm0.038}$} & \textbf{0.748$_{\pm0.067}$} & \textbf{0.824$_{\pm0.093}$} & \textbf{0.704$_{\pm0.086}$} \\
\midrule
\multirow{6}{*}{\rotatebox{90}{DSMIL}} 
& \multicolumn{3}{c|}{\textit{Baseline}}  & 0.487$_{\pm0.060}$ & 0.597$_{\pm0.048}$ & 0.634$_{\pm0.072}$ & 0.544$_{\pm0.010}$ & 0.729$_{\pm0.052}$ & 0.751$_{\pm0.035}$ & 0.679$_{\pm0.042}$ \\
&  &  & \Checkmark & 0.519$_{\pm0.035}$ & 0.653$_{\pm0.063}$ & 0.721$_{\pm0.034}$ & 0.563$_{\pm0.031}$ & 0.756$_{\pm0.036}$ & 0.782$_{\pm0.063}$ & 0.683$_{\pm0.033}$ \\
& \Checkmark &   &  & 0.527$_{\pm0.048}$ & 0.649$_{\pm0.057}$ & 0.685$_{\pm0.050}$ & 0.554$_{\pm0.029}$ & 0.733$_{\pm0.042}$ & 0.769$_{\pm0.047}$ & 0.673$_{\pm0.063}$ \\
& \Checkmark & \Checkmark  &  & \underline{0.539$_{\pm0.073}$} & \underline{0.692$_{\pm0.062}$} & \underline{0.728$_{\pm0.052}$} & \underline{0.585$_{\pm0.077}$} & \underline{0.793$_{\pm0.024}$} & 0.813$_{\pm0.083}$ & 0.699$_{\pm0.034}$ \\
& \Checkmark  &  & \Checkmark  & 0.532$_{\pm0.046}$ & 0.685$_{\pm0.044}$ & 0.719$_{\pm0.046}$ & 0.574$_{\pm0.068}$ & 0.782$_{\pm0.056}$ & \underline{0.852$_{\pm0.071}$} & \underline{0.708$_{\pm0.053}$} \\
& \Checkmark & \Checkmark & \Checkmark & \textbf{0.582$_{\pm0.084}$} & \textbf{0.729$_{\pm0.032}$} & \textbf{0.764$_{\pm0.029}$} & \textbf{0.620$_{\pm0.063}$} & \textbf{0.843$_{\pm0.082}$} & \textbf{0.868$_{\pm0.057}$} & \textbf{0.712$_{\pm0.027}$} \\
\midrule
\multirow{6}{*}{\rotatebox{90}{ABMIL}} 
& \multicolumn{3}{c|}{\textit{Baseline}}  & 0.502$_{\pm0.035}$ & 0.619$_{\pm0.046}$ & 0.647$_{\pm0.033}$ & 0.593$_{\pm0.052}$ & 0.806$_{\pm0.035}$ & 0.812$_{\pm0.040}$ & 0.682$_{\pm0.053}$ \\
&  &  & \Checkmark & 0.519$_{\pm0.061}$ & 0.653$_{\pm0.035}$ & 0.673$_{\pm0.058}$ & 0.608$_{\pm0.026}$ & 0.825$_{\pm0.062}$ & 0.831$_{\pm0.037}$ & 0.697$_{\pm0.045}$ \\
& \Checkmark &  &  & 0.514$_{\pm0.045}$ & 0.645$_{\pm0.028}$ & 0.685$_{\pm0.093}$ & 0.602$_{\pm0.073}$ & 0.814$_{\pm0.084}$ & 0.824$_{\pm0.068}$ & 0.693$_{\pm0.048}$ \\
& \Checkmark & \Checkmark  &  & 0.523$_{\pm0.061}$ & 0.664$_{\pm0.043}$ & 0.694$_{\pm0.082}$ & 0.613$_{\pm0.027}$ & 0.842$_{\pm0.036}$ & 0.835$_{\pm0.024}$ & 0.702$_{\pm0.064}$ \\
& \Checkmark  &  & \Checkmark  & \underline{0.527$_{\pm0.062}$} & \underline{0.678$_{\pm0.038}$} & \underline{0.705$_{\pm0.036}$} & \underline{0.624$_{\pm0.034}$} & \underline{0.856$_{\pm0.045}$} & \underline{0.883$_{\pm0.051}$} & \underline{0.709$_{\pm0.039}$} \\
& \Checkmark & \Checkmark & \Checkmark & \textbf{0.598$_{\pm0.087}$} & \textbf{0.729$_{\pm0.066}$} & \textbf{0.767$_{\pm0.034}$} & \textbf{0.665$_{\pm0.081}$} & \textbf{0.894$_{\pm0.072}$} & \textbf{0.920$_{\pm0.081}$} & \textbf{0.717$_{\pm0.028}$} \\
\bottomrule
\end{tabular}
}
\end{table*}

\subsection{Results and Discussion}
We evaluate CLCS on cancer subtype classification, IHC biomarker status classification, and survival prediction against:
1) MIL-based: TransMIL \cite{shao2021transmil}, DSMIL \cite{li2021dual}, ACMIL \cite{zhang2024attention}, FRMIL \cite{chikontwe2022feature}, and ABMIL \cite{ilse2018attention}; 2) foundation model-based: HIPT \cite{chen2022scaling}, GigaSSL \cite{lazard2023giga}, GigaPath \cite{xu2024whole}, and TANGLE \cite{jaume2024transcriptomics}; and 3) multi-stain methods: MADELEINE \cite{jaume2024multistain}. 

Our method outperforms SOTA approaches across subtasks with different $k$ values. 
For IHC biomarker status classification (Table~\ref{tab:comparison_IHCclassification}), CSCL exceeds all methods on BCNB and BRCA for $k\!\in\!\{1,5,10,25\}$. On BCNB, average AUCs are $0.598/0.681/0.729/0.767$ at $k\!=\!1/5/10/25$, surpassing MADELEINE by $+3.0\%/+0.3\%/+1.0\%/+0.2\%$, and on BRCA the gains are $+1.6\%/+3.3\%/+2.5\%/+0.1\%$. In survival prediction (Tables~\ref{tab:comparison_survivalprediction}), C-indexes reach $0.717$ (BRCA), $0.645$ (BLCA), $0.675$ (HNSC), and $0.768$ (COADREAD), yielding the highest average ($0.701$).
For subtype classification (Tables~\ref{tab:comparison_classification}), CSCL attains top AUCs: $0.665$ at $k{=}1$, $0.862$ at $k{=}5$, and $0.894$ at $k{=}10$. Gains arise from patch-level alignment, cross-stain fusion, and global alignment.

We ablate $\mathtt{CPA}$, $\mathtt{CAF}$, $\mathtt{CGA}$ on IHC biomarker status classification (BCNB), cancer subtype classification (BRCA), and survival prediction (BRCA) under TransMIL, DSMIL, and ABMIL (Table~\ref{tab:Ablation_CPA_CGA_CFA}).
The baseline removes $\mathtt{CPA}$, $\mathtt{CAF}$, and $\mathtt{CGA}$ from CSCL, using CONCH for patch-level features and MIL for slide-level representations, following MADELEINE training strategy~\cite{jaume2024multistain}. Baseline performance is lowest, adding $\mathtt{CPA}$ alone improves results. On TransMIL, BCNB IHC at $k=10$ increases from 0.535 to 0.541, BRCA subtype at $k=10$ from 0.623 to 0.640, and BRCA survival C-index from 0.668 to 0.673.
Similar gains on DSMIL and ABMIL indicate that $\mathtt{CPA}$ enforces patch-level alignment, with its adapter fine-tuning the encoder to inject IHC cues into H\&E.

Building on $\mathtt{CPA}$, $\mathtt{CAF}$ or $\mathtt{CGA}$ further boosts performance: on TransMIL, $\mathtt{CAF}$ lifts BCNB IHC at $k{=}10$ from 0.541 to 0.585, while $\mathtt{CGA}$ lifts it to 0.573. The same holds for subtype classification and survival prediction: $\mathtt{CAF}$ yields larger gains in low-shot classification (ABMIL BRCA, $k{=}10$: 0.814$\rightarrow$0.842) via cross-stain fusion, whereas $\mathtt{CGA}$ provides steadier improvements in survival prediction. Thus, $\mathtt{CAF}$ enriches representations via cross-stain feature fusion, whereas $\mathtt{CGA}$ aligns representations across stains, improving MIL for more informative feature extraction. Combining all modules yields the best results and consistent gains.

\begin{figure}[t]
    \centering
    \includegraphics[width=1\columnwidth]{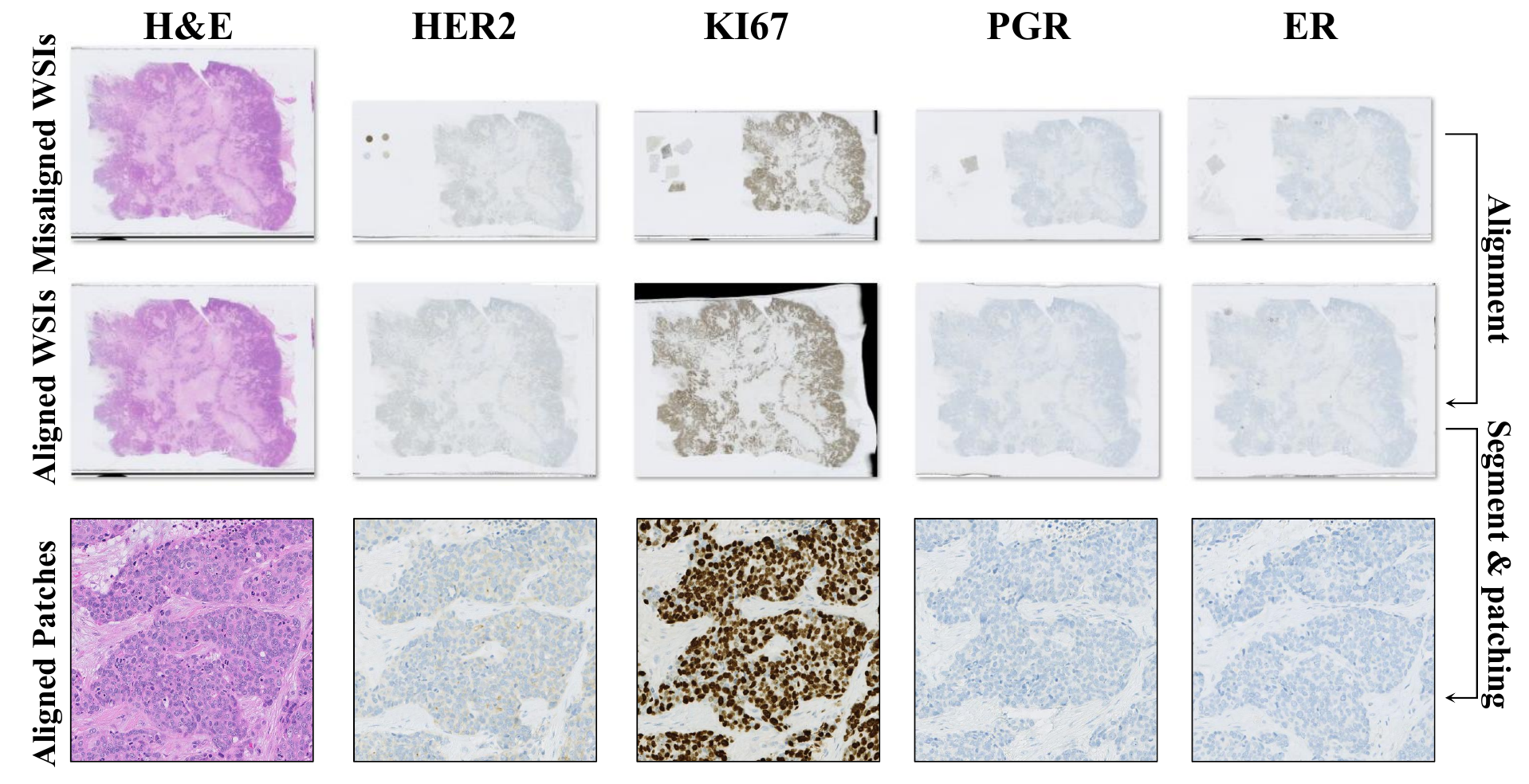} 
    \caption{Visual registration results for exemplary cases}
    \vspace{-2em}
    \label{fig:DATA}
\end{figure}

\begin{table}[t]
\centering
\caption{Results on two multi-stain paired datasets.}
\label{tab:datasets}
\resizebox{\columnwidth}{!}{
\begin{tabular}{c|c|c|c|c}
\toprule
Method & Dataset & ER ($\uparrow$) & PR ($\uparrow$) & HER2 ($\uparrow$) \\
\midrule
MADELEINE & ACROBAT & \underline{0.833\textsubscript{$\pm$0.038}} & 0.764\textsubscript{$\pm$0.051} & 0.699\textsubscript{$\pm$0.038} \\
& Our Dataset & 0.829\textsubscript{$\pm$0.021} & \underline{0.766\textsubscript{$\pm$0.037}} & \underline{0.702\textsubscript{$\pm$0.038}} \\
\midrule
CSCL (\textbf{Ours}) & ACROBAT & - & - & - \\
& Our Dataset & \textbf{0.841\textsubscript{$\pm$0.034}} & \textbf{0.775\textsubscript{$\pm$0.032}} & \textbf{0.685\textsubscript{$\pm$0.038}} \\
\bottomrule
\vspace{-4em}
\end{tabular}

}    
\end{table}

\noindent\textbf{Effectiveness of Our Dataset.}
Dataset effectiveness is assessed by comparing CSCL with MADELEINE on BCNB IHC biomarker classification under a 25-shot setting.
MADELEINE is trained on ACROBAT and our dataset, whereas CSCL is trained only on our dataset (ACROBAT lacks the patch-level alignment required by CSCL).
Specifically, MADELEINE is trained on ACROBAT and our aligned multi-stain dataset, whereas CSCL is trained only on our dataset (ACROBAT lacks the patch-level alignment CSCL requires).

As shown in Table~\ref{tab:datasets}, despite being $\sim$1/27 the size of ACROBAT, our dataset yields comparable MADELEINE AUC on ER (0.829 vs.\ 0.833) and slightly higher scores on PR (0.766 vs.\ 0.764) and HER2 (0.702 vs.\ 0.699).
CSCL trained solely on our dataset attains 0.841 (ER), 0.775 (PR), and 0.685 (HER2), surpassing MADELEINE on all biomarkers.

As in Fig.~\ref{fig:DATA}, our registration pipeline first aligns multi-stain WSIs at the slide level, then performs patch-level segmentation to ensure precise cross-stain correspondence.
These results show that well-aligned multi-stain data markedly improves performance and that CSCL effectively exploits this alignment.

\section{Conclusion}
Cross-Stain Contrastive Learning (CSCL) is a pretraining framework that leverages spatially aligned multi-stain WSIs to enhance slide-level representations. To address stain misalignment, a high-quality aligned dataset spanning H\&E and four IHC stains (HER2, KI67, ER, PGR) was curated. Extensive experiments show CSCL surpasses state-of-the-art in downstream tasks, yielding transferable, stain-invariant H\&E representations for real-world use.

\bibliographystyle{IEEEtran}
\bibliography{ref}


\end{document}